# An Analysis of Scale Invariance in Object Detection – SNIP

Bharat Singh    Larry S. Davis
University of Maryland, College Park
{bharat,lsd}@cs.umd.edu

## Abstract

*An analysis of different techniques for recognizing and detecting objects under extreme scale variation is presented. Scale specific and scale invariant design of detectors are compared by training them with different configurations of input data. By evaluating the performance of different network architectures for classifying small objects on ImageNet, we show that CNNs are not robust to changes in scale. Based on this analysis, we propose to train and test detectors on the same scales of an image-pyramid. Since small and large objects are difficult to recognize at smaller and larger scales respectively, we present a novel training scheme called Scale Normalization for Image Pyramids (SNIP) which selectively back-propagates the gradients of object instances of different sizes as a function of the image scale. On the COCO dataset, our single model performance is 45.7% and an ensemble of 3 networks obtains an mAP of 48.3%. We use off-the-shelf ImageNet-1000 pre-trained models and only train with bounding box supervision. Our submission won the Best Student Entry in the COCO 2017 challenge. Code will be made available at* http://bit.ly/2yXVg4c.

## 1. Introduction

Deep learning has fundamentally changed how computers perform image classification and object detection. In less than five years, since AlexNet [20] was proposed, the top-5 error on ImageNet classification [9] has dropped from 15% to 2% [16]. This is super-human level performance for image classification with 1000 classes. On the other hand, the mAP of the best performing detector [18] (which is only trained to detect 80 classes) on COCO [25] is only 62% – even at 50% overlap. Why is object detection so much harder than image classification?

Large scale variation across object instances, and especially, the challenge of detecting very small objects stands out as one of the factors behind the difference in performance. Interestingly, the median scale of object instances relative to the image in ImageNet (classification) vs COCO (detection) are 0.554 and 0.106 respectively. Therefore, most object instances in COCO are smaller than 1% of image area! To make matters worse, the scale of the smallest and largest 10% of object instances in COCO is 0.024 and 0.472 respectively (resulting in scale variations of almost 20 times!); see Fig. 1. This variation in scale which a detector needs to handle is enormous and presents an extreme challenge to the scale invariance properties of convolutional neural networks. Moreover, differences in the scale of object instances between classification and detection datasets also results in a large *domain-shift* while fine-tuning from a pre-trained classification network. In this paper, we first provide evidence of these problems and then propose a training scheme called Scale Normalization for Image Pyramids which leads to a state-of-the-art object detector on COCO.

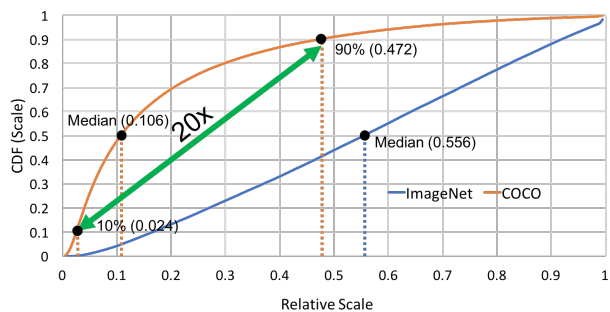

Figure 1. Fraction of RoIs in the dataset vs scale of RoIs relative to the image.

To alleviate the problems arising from scale variation and small object instances, multiple solutions have been proposed. For example, features from the layers near to the input, referred to as shallow(er) layers, are combined with deeper layers for detecting small object instances [23, 34, 1, 13, 27], dilated/deformable convolution is used to increase receptive fields for detecting large objects [32, 7, 37, 8], independent predictions at layers of different resolutions are used to capture object instances of different scales [36, 3, 22], context is employed for disambiguation [38, 39, 10], training is performed over a range of scales [7, 8, 15] or, inference is performed on multiple scales of an image pyramid and predictions are combined using non-maximum suppression [7, 8, 2, 33].

While these architectural innovations have significantly helped to improve object detection, many important issues related to training remain unaddressed:

- Is it critical to upsample images for obtaining good performance for object detection? Even though the typical size of images in detection datasets is 480x640, why is it a common practice to up-sample them to 800x1200? Can we pre-train CNNs with smaller strides on low resolution images from ImageNet and then fine-tune them on detection datasets for detecting small object instances?

- When fine-tuning an object detector from a pre-trained image classification model, should the resolution of the training object instances be restricted to a tight range (from 64x64 to 256x256) after appropriately re-scaling the input images, or should all object resolutions (from 16x16 to 800x1000, in the case of COCO) participate in training after up-sampling input images?

We design controlled experiments on ImageNet and COCO to seek answers to these questions. In Section 3, we study the effect of scale variation by examining the performance of existing networks for ImageNet classification when images of different scales are provided as input. We also make minor modifications to the CNN architecture for classifying images of different scales. These experiments reveal the importance of up-sampling for small object detection. To analyze the effect of scale variation on object detection, we train and compare the performance of scale-specific and scale invariant detector designs in Section 5. For scale-specific detectors, variation in scale is handled by training separate detectors - one for each scale range. Moreover, training the detector on similar scale object instances as the pre-trained classification networks helps to reduce the domain shift for the pre-trained classification network. But, scale-specific designs also reduce the number of training samples per scale, which degrades performance. On the other hand, training a single object detector with all training samples makes the learning task significantly harder because the network needs to learn filters for detecting object instances over a wide range of scales.

Based on these observations, in Section 6 we present a novel training paradigm, which we refer to as Scale Normalization for Image Pyramids (SNIP), that benefits from reducing scale-variation during training but without paying the penalty of reduced training samples. Scale-invariance is achieved using an image-pyramid (instead of a scale-invariant detector), which contains normalized input representations of object instances in one of the scales in the image-pyramid. To minimize the domain shift for the classification network during training, we only back-propagate gradients for RoIs/anchors that have a resolution close to that of the pre-trained CNN. Since we train on each scale in the pyramid with the above constraint, SNIP effectively utilizes all the object instances available during training. The proposed approach is generic and can be plugged into the training pipeline of different problems like instance-segmentation, pose-estimation, spatio-temporal action detection - wherever the "objects" of interest manifest large scale variations.

Contrary to the popular belief that deep neural networks can learn to cope with large variations in scale given enough training data, we show that SNIP offers significant improvements (3.5%) over traditional object detection training paradigms. Our ensemble with a Deformable-RFCN backbone obtains an mAP of 69.7% at 50% overlap, which is an improvement of 7.4% over the state-of-the-art on the COCO dataset.

## 2. Related Work

Scale space theory [35, 26] advocates learning representations that are invariant to scale and the theory has been applied to many problems in the history of computer vision [4, 30, 28, 21, 14, 5, 23]. For problems like object detection, pose-estimation, instance segmentation etc., learning scale invariant representations is critical for recognizing and localizing objects. To detect objects at multiple scales, many solutions have been proposed.

The deeper layers of modern CNNs have large strides (32 pixels) that lead to a very coarse representation of the input image, which makes small object detection very challenging. To address this problem, modern object detectors [32, 7, 5] employ dilated/atrous convolutions to increase the resolution of the feature map. Dilated/deformable convolutions also preserve the weights and receptive fields of the pre-trained network and do not suffer from degraded performance on large objects. Up-sampling the image by a factor of 1.5 to 2 times during training and up to 4 times during inference is also a common practice to increase the final feature map resolution [8, 7, 15]. Since feature maps of layers closer to the input are of higher resolution and often contain complementary information (wrt. conv5), these features are either combined with shallower layers (like conv4, conv3) [23, 31, 1, 31] or independent predictions are made at layers of different resolutions [36, 27, 3]. Methods like SDP [36], SSH [29] or MS-CNN [3], which make independent predictions at different layers, also ensure that smaller objects are trained on higher resolution layers (like conv3) while larger objects are trained on lower resolution layers (like conv5). This approach offers better resolution at the cost of high-level semantic features which can hurt performance.

Methods like FPN, Mask-RCNN, RetinaNet [23, 13, 24], which use a pyramidal representation and combine features of shallow layers with deeper layers at least have access to higher level semantic information. However, if the size of an object was 25x25 pixels then even an up-sampling factor of 2 during training,will scale the object to only 50x50 pixels. Note that typically the network is pre-trained on images of resolution 224x224. Therefore, the high level semantic features (at conv5) generated even by feature pyramid networks will not be useful for classifying small objects (a similar argument can be made for large objects in high resolution images). Hence, combining them with features from shallow layers would not be good for detecting small objects, see Fig. 2. Although feature pyramids efficiently exploit features from all the layers in the network, they are not an attractive alternative to an image pyramid for detecting very small/large objects.

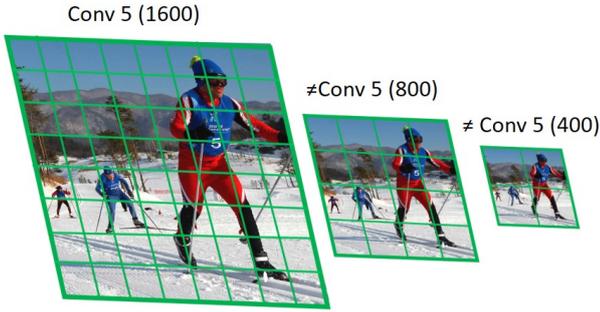

Figure 2. The same layer convolutional features at different scales of the image are different and map to different semantic regions in the image at different scales.

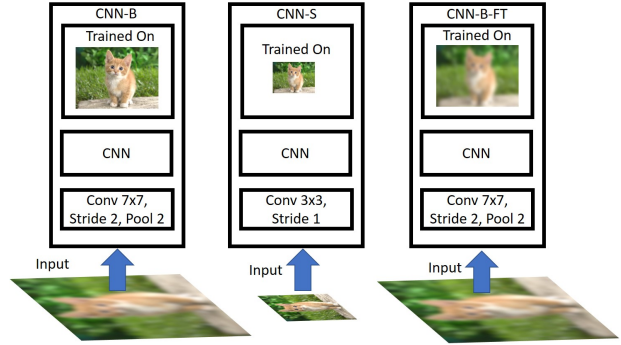

Figure 3. Both CNN-B and CNN-B-FT are provided an upsampled low resolution image as input. CNN-S is provided a low resolution image as input. CNN-B is trained on high resolution images. CNN-S is trained on low resolution images. CNN-B-FT is pre-trained on high resolution images and fine-tuned on upsampled low-resolution images. ResNet-101 architecture is used.

Recently, a pyramidal approach was proposed for detecting faces [17] where the gradients of all objects were back-propagated after max-pooling the responses from each scale. Different filters were used in the classification layers for faces at different scales. This approach has limitations for object detection because training data per class in object detection is limited and the variations in appearance, pose etc. are much larger compared to face detection. We observe that adding scale specific filters in R-FCN for each class hurts performance for object detection. In [33], an image pyramid was generated and maxout [12] was used to select features from a pair of scales closer to the resolution of the pre-trained dataset during inference. A similar inference procedure was also proposed in SPPNet and Fast-RCNN [14, 11]: however, standard multi-scale training (described in Section 5) was used. We explore the design space for *training* scale invariant object detectors and propose to selectively back-propagate gradients for samples close to the resolution of the pre-trained network.

## 3. Image Classification at Multiple Scales

In this section we study the effect of domain shift, which is introduced when different resolutions of images are provided as input during training and testing. We perform this analysis because state-of-the-art detectors are typically trained at a resolution of 800x1200 pixels [1], but inference is performed on an image pyramid, including higher resolutions like 1400x2000 for detecting small objects [8, 7, 2].

**Naïve Multi-Scale Inference:** Firstly, we obtain images at different resolutions, 48x48, 64x64, 80x80, 96x96 and 128x128, by down-sampling the original ImageNet database. These are then up-sampled to 224x224 and provided as input to a CNN architecture trained on 224x224 size images, referred to as CNN-B (see Fig. 3). Fig. 4 (a) shows the top-1 accuracy of CNN-B with a ResNet-101 backbone. We observe that as the difference in resolution between training and testing images increases, so does the drop in performance. Hence, testing on resolutions on which the network was not trained is clearly sub-optimal, at least for image classification.

**Resolution Specific Classifiers:** Based on the above observation, a simple solution for improving the performance of detectors on smaller objects is to pre-train classification networks with a different stride on ImageNet. After-all, network architectures which obtain best performance on CIFAR10 [19] (which contains small objects) are different from ImageNet. The first convolution layer in ImageNet classification networks has a stride of 2 followed by a max pooling layer of stride 2x2, which can potentially wipe out most of the image signal present in a small object. Therefore, we train ResNet-101 with a stride of 1 and 3x3 convolutions in the first layer for 48x48 images (CNN-S, see

---

[1] original image resolution is typically 480x640

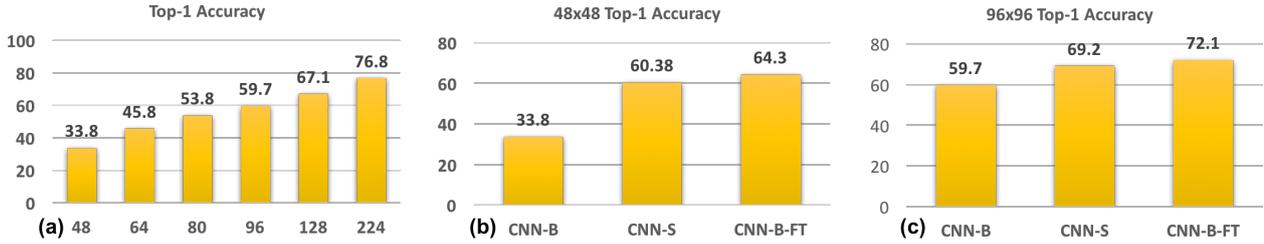

Figure 4. All figures report accuracy on the validation set of the ImageNet classification dataset. We upsample images of resolution 48,64,80 etc. and plot the Top-1 accuracy of the pre-trained ResNet-101 classifier in figure (a). Figure (b,c) show results for different CNNs when the original image resolution is 48,96 pixels respectively.

Fig. 3), a typical architecture used for CIFAR. Similarly, for 96x96 size images, we use a kernel of size 5x5 and stride of 2. Standard data augmentation techniques such as random cropping, color augmentation, disabling color augmentation after 70 epochs are used to train these networks. As seen in Fig. 4, these networks (CNN-S) perform significantly better than CNN-B. Therefore, it is tempting to pre-train classification networks with different architectures for low resolution images and use them for object detection for low resolution objects.

**Fine-tuning High-Resolution Classifiers:** Yet another simple solution for small object detection would be to fine-tune CNN-B on up-sampled low resolution images to yield CNN-B-FT ( Fig. 3). The performance of CNN-B-FT on up-sampled low-resolution images is better than CNN-S, Fig. 4. This result empirically demonstrates that the filters learned on high-resolution images can be useful for recognizing low-resolution images as well. Therefore, instead of reducing the stride by 2, it is better to up-sample images 2 times and then fine-tune the network pre-trained on high-resolution images.

While training object detectors, we can either use different network architectures for classifying objects of different resolutions or use the a single architecture for all resolutions. Since pre-training on ImageNet (or other larger classification datasets) is beneficial and filters learned on larger object instances help to classify smaller object instances, upsampling images and using the network pre-trained on high resolution images should be better than a specialized network for classifying small objects. Fortunately, existing object detectors up-sample images for detecting smaller objects instead of using a different architecture. Our analysis supports this practice and compares it with other alternatives to emphasize the difference.

## 4. Background

In the next section, we discuss a few baselines for detecting small objects. We briefly describe the Deformable-RFCN [8] detector which will be used in the following analysis. D-RFCN obtains the best single model results on COCO and is publicly available, so we use this detector.

Deformable-RFCN is based on the R-FCN detector [7]. It adds deformable convolutions in the conv5 layers to adaptively change the receptive field of the network for creating scale invariant representations for objects of different scales. At each convolutional feature map, a lightweight network predicts offsets on the 2D grid, which are spatial locations at which spatial sub-filters of the convolution kernel are applied. The second change is in Position Sensitive RoI Pooling. Instead of pooling from a fixed set of bins on the convolutional feature map (for an RoI), a network predicts offsets for each position sensitive filter (depending on the feature map) on which Position Sensitive RoI (PSRoI)-Pooling is performed.

For our experiments, proposals are extracted at a single resolution (after upsampling) of 800x1200 using a publicly available Deformable-RFCN detector. It has a ResNet-101 backbone and is trained at a resolution of 800x1200. 5 anchor scales are used in RPN for generating proposals [2]. For classifying these proposals, we use Deformable-RFCN with a ResNet-50 backbone without the Deformable Position Sensitive RoIPooling. We use Position Sensitive RoIPooling with bilinear interpolation as it reduces the number of filters by a factor of 3 in the last layer. NMS with a threshold of 0.3 is used. Not performing end-to-end training along with RPN, using ResNet-50 and eliminating deformable PSRoI filters reduces training time by a factor of 3 and also saves GPU memory.

## 5. Data Variation or Correct Scale?

The study in section 3 confirms that differences in resolutions between the training and testing phase leads to a significant drop in performance. Unfortunately, this difference in resolution is part of the current object detection pipeline - due to GPU memory constraints, training is performed on a lower resolution (800x1200) than testing (1400x2000) (note that original resolution is typically 640x480). This section analyses the effect of image resolution, the scale of object instances and variation in data on the performance of an object detector. We train detectors under different settings and evaluate them on 1400x2000 images for detecting small objects (less than 32x32 pixels in the COCO dataset) only to tease apart the factors that affect the performance. The results are reported in Table 1.

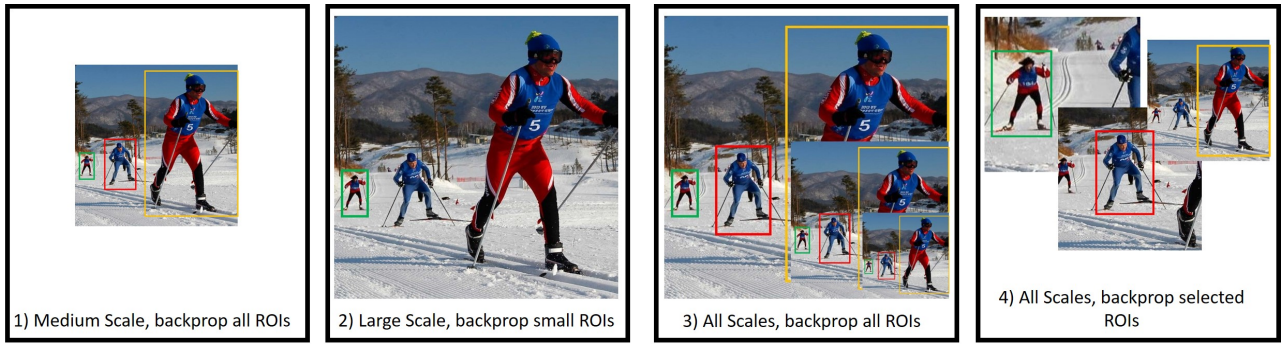

Figure 5. Different approaches for providing input for training the classifier of a proposal based detector.

**Training at different resolutions:** We start by training detectors that use all the object instances on two different resolutions, 800x1400 and 1400x2000, referred to as $800_{all}$ and $1400_{all}$, respectively, Fig 5.1. As expected, $1400_{all}$ outperformed $800_{all}$, because the former is trained and tested on the same resolution i.e. 1400x2000. However, the improvement is only marginal. Why? To answer this question we consider what happens to the medium-to-large object instances while training at such a large resolution. They become too big to be correctly classified! Therefore, training at higher resolutions scales up small objects for better classification, but blows up the medium-to-large objects which degrades performance.

**Scale specific detectors:** We trained another detector ($1400_{<80px}$) at a resolution of 1400x2000 while ignoring all the medium-to-large objects ($> 80$ pixels, in the original image) to eliminate the deleterious-effects of extremely large objects, Fig 5.2. Unfortunately, it performed significantly worse than even $800_{all}$. What happened? We lost a significant source of variation in appearance and pose by ignoring medium-to-large objects (about 30% of the total object instances) that hurt performance more than it helped by eliminating extreme scale objects.

**Multi-Scale Training (MST):** Lastly, we evaluated the common practice of obtaining scale-invariant detectors by using randomly sampled images at multiple resolutions during training, referred to as MST [2], Fig 5.3. It ensures training instances are observed at many different resolutions, but it also degraded by extremely small and large objects. It performed similar to $800_{all}$. We conclude that it is important to train a detector with appropriately scaled objects while capturing as much variation across the objects as possible. In the next section we describe our proposed solution that achieves exactly this and show that it outperforms current

[2]MST also uses a resolution of 480x800

| $1400_{<80px}$ | $800_{all}$ | $1400_{all}$ | MST | SNIP |
|---|---|---|---|---|
| 16.4 | 19.6 | 19.9 | 19.5 | 21.4 |

Table 1. mAP on Small Objects (smaller than 32x32 pixels) under different training protocols. MST denotes multi-scale training as shown in Fig. 5.3. R-FCN detector with ResNet-50 (see Section 4).

training pipelines.

## 6. Object Detection on an Image Pyramid

Our goal is to combine the best of both approaches i.e. train with maximal variations in appearance and pose while restricting scale to a reasonable range. We achieve this by a novel construct that we refer to as Scale Normalization for Image Pyramids (SNIP). We also discuss details of training object detectors on an image pyramid within the memory limits of current GPUs.

### 6.1. Scale Normalization for Image Pyramids

SNIP is a modified version of MST where only the object instances that have a resolution close to the pre-training dataset, which is typically 224x224, are used for training the detector. In multi-scale training (MST), each image is observed at different resolutions therefore, at a high resolution (like 1400x2000) large objects are hard to classify and at a low resolution (like 480x800) small objects are hard to classify. Fortunately, each object instance appears at several different scales and some of those appearances fall in the desired scale range. In order to eliminate extreme scale objects, either too large or too small, training is only performed on objects that fall in the desired scale range and the remainder are simply ignored during back-propagation. Effectively, SNIP uses all the object instances during training, which helps capture all the variations in appearance and pose, while reducing the *domain-shift* in the scale-space for the pre-trained network. The result of evaluating the detector trained using SNIP is reported in Table 1 - it outperforms all the other approaches. This experiment demonstrates the effectiveness of SNIP for detecting small objects. Below we discuss the implementation of SNIP in detail.

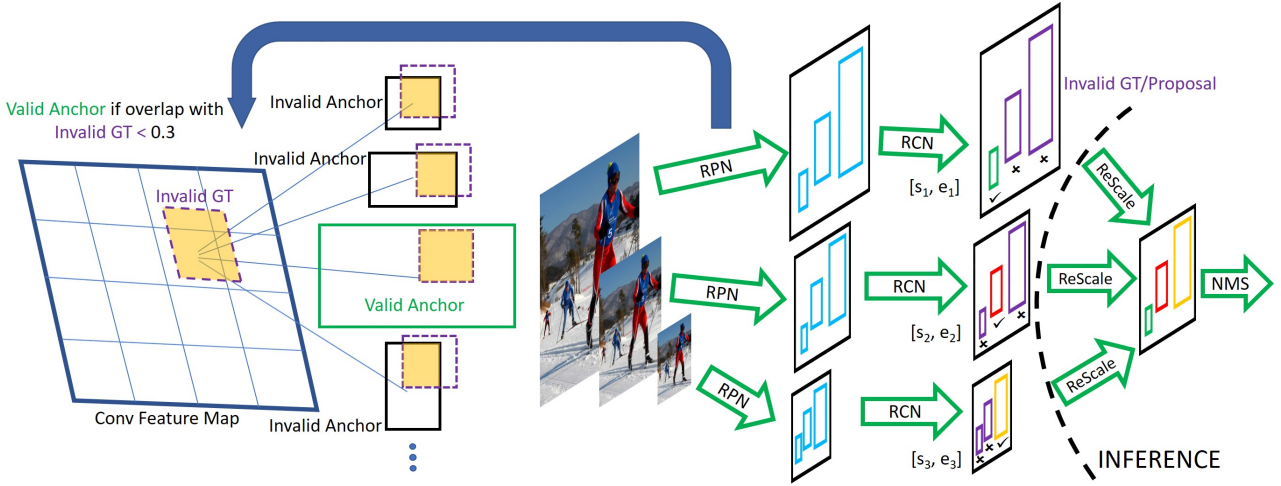

Figure 6. SNIP training and inference is shown. Invalid RoIs which fall outside the specified range at each scale are shown in purple. These are discarded during training and inference. Each batch during training consists of images sampled from a particular scale. Invalid GT boxes are used to invalidate anchors in RPN. Detections from each scale are rescaled and combined using NMS.

For training the classifier, all ground truth boxes are used to assign labels to proposals. We do not select proposals and ground truth boxes which are outside a specified size range at a particular resolution during training. At a particular resolution $i$, if the area of an RoI $ar(r)$ falls within a range $[s_i^c, e_i^c]$, it is marked as valid, else it is invalid. Similarly, RPN training also uses all ground truth boxes to assign labels to anchors. Finally, those anchors which have an overlap greater than 0.3 with an invalid ground truth box are excluded during training (i.e. their gradients are set to zero). During inference, we generate proposals using RPN for each resolution and classify them independently at each resolution as shown in Fig 6. Similar to training, we do not select detections (not proposals) which fall outside a specified range at each resolution. After classification and bounding-box regression, we use soft-NMS [2] to combine detections from multiple resolutions to obtain the final detection boxes, refer to Fig. 6.

The resolution of the RoIs after pooling matches the pre-trained network, so it is easier for the network to learn during fine-tuning. For methods like R-FCN which divide RoIs into sub-parts and use position sensitive filters, this becomes even more important. For example, if the size of an RoI is 48 pixels (3 pixels in the conv5 feature map) and there are 7 filters along each axis, the positional correspondence between features and filters would be lost.

### 6.2. Sampling Sub-Images

Training on high resolution images with deep networks like ResNet-101 or DPN-92 [6] requires more GPU memory. Therefore, we crop images so that they fit in GPU memory. Our aim is to generate the minimum number of chips (sub-images) of size 1000x1000 which cover all the small objects in the image. This helps in accelerating training as no computation is needed where there are no small objects. For this, we generate 50 randomly positioned chips of size 1000x1000 per image. The chip which covers the maximum number of objects is selected and added to our set of training images. Until all objects in the image are covered, we repeat the sampling and selection process on the remaining objects. Since chips are randomly generated and proposal boxes often have a side on the image boundary, for speeding up the sampling process we snap the chips to image boundaries. We found that, on average, 1.7 chips of size 1000x1000 are generated for images of size 1400x2000. This sampling step is not needed when the image resolution is 800x1200 or 480x640 or when an image does not contain small objects. Random cropping is not the reason why we observe an improvement in performance for our detector. To verify this, we trained ResNet-50 (as it requires less memory) using un-cropped high-resolution images (1400x2000) and did not observe any change in mAP.

## 7. Datasets and Evaluation

We evaluate our method on the COCO dataset. COCO contains 123,000 images for training and evaluation is performed on 20,288 images in test-dev. Since recall for proposals is not provided by the evaluation server on COCO, we train on 118,000 images and report recall on the remaining 5,000 images (commonly referred to as minival set). Unless specifically mentioned, the area of small objects is less than 32x32, medium objects range from 32x32 to 96x96 and large objects are greater than 96x96.

| Method | AP | $AP^S$ | $AP^M$ | $AP^L$ |
|---|---|---|---|---|
| Single scale | 34.5 | 16.3 | 37.2 | 47.6 |
| MS Test | 35.9 | 19.5 | 37.3 | 48.5 |
| MS Train/Test | 35.6 | 19.5 | 37.5 | 47.3 |
| SNIP | 37.8 | 21.4 | 40.4 | 50.1 |

Table 2. MS denotes multi-scale. Single scale is (800,1200). R-FCN detector with ResNet-50 (as described in Section 4).

| Method | AR | $AR^{50}$ | $AR^{75}$ | 0-25 | 25-50 | 50-100 |
|---|---|---|---|---|---|---|
| Baseline | 57.6 | 88.7 | 67.9 | 67.5 | 90.1 | 95.6 |
| + Improved | 61.3 | 89.2 | 69.8 | 68.1 | 91.0 | 96.7 |
| + SNIP | 64.0 | 92.1 | 74.7 | 74.4 | 95.1 | 98.0 |
| DPN-92 | 65.7 | 92.8 | 76.3 | 76.7 | 95.7 | 98.2 |

Table 3. For individual ranges (like 0-25 etc.) recall at 50% overlap is reported because minor localization errors can be fixed in the second stage. First three rows use ResNet-50 as the backbone. Recall is for 900 proposals, as top 300 are taken from each scale.

## 7.1. Training Details

We train Deformable-RFCN [8] as our detector with 3 resolutions, (480, 800), (800, 1200) and (1400,2000), where the first value is for the shorter side of the image and the second one is the limit on the maximum size of a side. Training is performed for 7 epochs for the classifier while RPN is trained for 6 epochs. Although it is possible to combine RPN and RCN using alternating training which leads to slight improvement in accuracy [23], we train separate models for RPN and RCN and evaluate their performance independently. This is because it is faster to experiment with different classification architectures after proposals are extracted. We use a warmup learning rate of 0.0005 for 1000 iterations after which it is increased to 0.005. Step down is performed at 4.33 epochs for RPN and 5.33 epochs otherwise. For our baselines which did not involve SNIP, we also evaluated their performance after 8 or 9 epochs but observed that results after 7 epochs were best. For the classifier (RCN), on images of resolution (1400,2000), the valid range in the original image (without up/down sampling) is [0, 80], at a resolution of (800,1200) it is [40, 160] and at a resolution of (480,800) it is [120, $\infty$]. We have an overlap of 40 pixels over adjacent ranges. These ranges were design decisions made during training, based on the consideration that after re-scaling, the resolution of the valid RoIs does not significantly differ from the resolution on which the backbone CNN was trained. Since in RPN even a one pixel feature map can generate a proposal we use a validity range of [0,160] at (800,1200) for valid ground truths for RPN. For inference, the validity range for each resolution in RCN is obtained using the minival set. Training RPN is fast so we enable SNIP after the first epoch. SNIP doubles the training time per epoch, so we enable it after 3 epochs for training RCN.

## 7.2. Improving RPN

In detectors like Faster-RCNN/R-FCN, Deformable R-FCN, RPN is used for generating region proposals. RPN assigns an anchor as positive only if overlap with a ground truth bounding box is greater than 0.7 [3]. We found that when using RPN at conv4 with 15 anchors (5 scales - 32, 64, 128, 256, 512, stride 16, 3 aspect ratios), only 30% of the ground truth boxes match this criterion when the image resolution is 800x1200 in COCO. Even if this threshold is changed to 0.5, only 58% of the ground truth boxes have an anchor which matches this criterion. Therefore, for more than 40% of the ground truth boxes, an anchor which has an overlap less than 0.5 is assigned as a positive (or ignored). Since we sample the image at multiple resolutions and back-propagate gradients at the relevant resolution only, this problem is alleviated to some extent. We also concatenate the output of conv4 and conv5 to capture diverse features and use 7 anchor scales. A more careful combination of features with predictions at multiple layers like [23, 13] should provide a further boost in performance.

## 7.3. Experiments

First, we evaluate the performance of SNIP on classification (RCN) under the same settings as described in Section 4. In Table 2, performance of the single scale model, multi-scale testing, and multi-scale training followed by multi-scale testing is shown. We use the best possible validity ranges at each resolution for each of these methods when multi-scale testing is performed. Multi-scale testing improves performance by 1.4%. Performance of the detector deteriorates for large objects when we add multi-scale training. This is because at extreme resolutions the receptive field of the network is not sufficient to classify them. SNIP improves performance by 1.9% compared to standard multi-scale testing. Note that we only use single scale proposals common across all three scales during classification for this experiment.

For RPN, a baseline with the ResNet-50 network was trained on the conv4 feature map. Top 300 proposals are selected from each scale and all these 900 proposals are used for computing recall. Average recall (averaged over multiple overlap thresholds, just like mAP) is better for our improved RPN, as seen in Table 3. This is because for large objects ($> 100$ pixels), average recall improves by 10% (not shown in table) for the improved baseline. Although the improved version improves average recall, it does not have much effect at 50% overlap. Recall at 50% is most important for object proposals because bounding box regression can correct minor localization errors, but if an object is not covered at all by proposals, it will clearly lead to a miss. Recall for objects greater than 100 pixels at 50% overlap is already close to 100%, so improving average recall for large objects is not that valuable for a detector. Note that SNIP improves recall at 50% overlap by 2.9% compared to our improved baseline. For objects smaller than 25 pixels, the improvement in recall is 6.3%. Using a stronger classification network like DPN-92 also improves recall. In last two rows of Table 4, we perform an ablation study with our best model, which uses a DPN-98 classifier and DPN-92 proposals. If we train our improved RPN without SNIP, mAP drops by 1.1% on small objects and 0.5% overall. Note that AP of large objects is not affected as we still use the classification model trained with SNIP.

---
[3] If there does not exist a matching anchor, RPN assigns the anchor with the maximum overlap with ground truth bounding box as positive.

| Method | Backbone | AP | $AP^{50}$ | $AP^{75}$ | $AP^S$ | $AP^M$ | $AP^L$ |
|---|---|---|---|---|---|---|---|
| D-RFCN [8, 2] | ResNet-101 | 38.4 | 60.1 | 41.6 | 18.5 | 41.6 | 52.5 |
| Mask-RCNN [13] | ResNext-101 (seg) | 39.8 | 62.3 | 43.4 | 22.1 | 43.2 | 51.2 |
| D-RFCN [8, 2] | ResNet-101 (6 scales) | 40.9 | 62.8 | 45.0 | 23.3 | 43.6 | 53.3 |
| G-RMI [18] | Ensemble | 41.6 | 62.3 | 45.6 | 24.0 | 43.9 | 55.2 |
| D-RFCN | DPN-98 | 41.2 | 63.5 | 45.9 | 25.7 | 43.9 | 52.8 |
| D-RFCN + SNIP (RCN) | DPN-98 | 44.2 | 65.6 | 49.7 | 27.4 | 47.8 | 55.8 |
| D-RFCN + SNIP (RCN+RPN) | DPN-98 | 44.7 | 66.6 | 50.2 | 28.5 | 47.8 | 55.9 |
| Faster-RCNN + SNIP (RPN) | ResNet-101 | 43.1 | 65.3 | 48.1 | 26.1 | 45.9 | 55.2 |
| Faster-RCNN + SNIP (RPN+RCN) | ResNet-101 | 44.4 | 66.2 | 49.9 | 27.3 | 47.4 | 56.9 |
| D-RFCN + SNIP | ResNet-101 (ResNet-101 proposals) | 43.4 | 65.5 | 48.4 | 27.2 | 46.5 | 54.9 |
| | DPN-98 (with flip) | 45.7 | 67.3 | 51.1 | 29.3 | 48.8 | 57.1 |
| | Ensemble | **48.3** | **69.7** | **53.7** | **31.4** | **51.6** | **60.7** |

Table 4. Comparison with state-of-the-art detectors. (seg) denotes that segmentation masks were also used. We train on train+val and evaluate on test-dev. Unless mentioned, we use 3 scales and DPN-92 proposals. Ablation for SNIP in RPN and RCN is shown.

Finally, we compare with state-of-the-art detectors in Table 4. For these experiments, we use the deformable position sensitive filters and Soft-NMS. Compared to the single scale deformable R-FCN baseline shown in the first line of Table 4, multi-scale training and inference improves overall results by 5% and for small objects by 8.7%! This shows the importance of an image pyramid for object detection. Compared to the best single model method (which uses 6 instead of 3 scales and is also trained end-to-end) based on ResNet-101, we improve performance by 2.5% overall and 3.9% for small objects. We observe that using better backbone architectures further improves the performance of the detector. When SNIP is not used for both the proposals and the classifier, mAP drops by 3.5% for the DPN-98 classifier, as shown in the last row. For the ensemble, DPN-92 proposals are used for all the networks (including ResNet-101). Since proposals are shared across all networks, we average the scores and box-predictions for each RoI. During flipping we average the detection scores and bounding box predictions. Finally, Soft-NMS is used to obtain the final detections. Iterative bounding-box regression is not used. All pre-trained models are trained on ImageNet-1000 and COCO segmentation masks are not used. Faster-RCNN was not used in the ensemble. On 100 images, it takes 90 seconds for to perform detection on a Titan X GPU using a ResNet-101 backbone. Speed can be improved with end-to-end training (we perform inference for RPN and RCN separately).

We also conducted experiments with the Faster-RCNN detector with deformable convolutions. Since the detector does not have position-sensitive filters, it is more robust to scale and performs better for large objects. Training it with SNIP still improves performance by 1.3%. Note that we can get an mAP of 44.4% with a single head faster-RCNN without using any feature-pyramid!

## 8. Conclusion

We presented an analysis of different techniques for recognizing and detecting objects under extreme scale variation, which exposed shortcomings of the current object detection training pipeline. Based on the analysis, a training scheme (SNIP) was proposed to tackle the wide scale spectrum of object instances which participate in training and to reduce the domain-shift for the pre-trained classification network. Experimental results on the COCO dataset demonstrated the importance of scale and image-pyramids in object detection. Since we do not need to back-propagate gradients for large objects in high-resolution images, it is possible to reduce the computation performed in a significant portion of the image. We would like to explore this direction in future work.

**Acknowledgement** We would like to thank Abhishek Sharma for helpful discussions and for improving the presentation of the paper. The research was supported by the Office of Naval Research under Grant N000141612713: Visual Common Sense Reasoning for Multi-agent Activity Prediction and Recognition.